\documentclass{elsarticle}

\usepackage{amsmath}
\usepackage{amssymb}
\usepackage{mathtools}
\usepackage{multirow}
\usepackage{stmaryrd}
\usepackage{multirow}
\usepackage{etoolbox}
\usepackage[hyphens]{url}
\usepackage[breaklinks]{hyperref}
\usepackage{graphicx}
\usepackage{booktabs} 
\usepackage{rotating}
\usepackage[bottom]{footmisc}

\usepackage{float}
\usepackage[caption = false]{subfig}
\usepackage{placeins}

\def\ptFiguresDirectory#1{./figures/#1}
\def\ivBrack#1{\left\llbracket{#1}\right\rrbracket} 
\def\FWER#1{FWER}
\def\fdr{\mathrm{FDR}}
\def\fnr{\mathrm{FNR}}
\def\fOne{{F_{1}}}
\def\neighCoeff{\zeta}
\def\dotProd#1#2{\left\langle {#1};{#2}\right\rangle}
\def\vnorm#1{\left\lVert {#1} \right\rVert}
\DeclareMathOperator{\sign}{sign}
\def\setBracks#1{\left\{{#1}\right\}}
\def\classMahDist{dc}
\def\classNormDist{dn}
\def\projection#1#2{\mathrm{proj}\left({#1},{#2} \right)}
\def\classCovar{S}
\def\classProjCovar{S\!n}
\def\ivBrack#1{\left\llbracket{#1}\right\rrbracket}

\journal{DSS}

\begin{document}

\begin{frontmatter}


\title{Linear Classifier Combination via Multiple Potential Functions}
\author{Pawel Trajdos\corref{cor1}}
\author{Robert Burduk}

\cortext[cor1]{pawel.trajdos@pwr.edu.pl}
\address{Department of Systems and Computer Networks, Wroclaw
University of Science and Technology, \\ Wybrzeze Wyspianskiego 27, 50-370
Wroclaw, Poland }

\begin{abstract}

A vital aspect of the classification based model construction process is the calibration of the scoring function. One of the weaknesses of the calibration process is that it does not take into account the information about the relative positions of the recognized objects in the feature space. To alleviate this limitation, in this paper, we propose a novel concept of calculating a scoring function based on the distance of the object from the decision boundary and its distance to the class centroid. An important property is that the proposed score function has the same nature for all linear base classifiers, which means that outputs of these classifiers are equally represented and have the same meaning. The proposed approach is compared with other ensemble algorithms and experiments on multiple Keel datasets demonstrate the effectiveness of our method. To discuss the results of our experiments, we use multiple classification performance measures and statistical analysis.

\end{abstract}


\end{frontmatter}

\section{Introduction}\label{sec:introduction}





%

Ensemble methods are a popular approach to improving the possibilities of individual supervised classification algorithms (base learners, base classifiers) by building more stable and accurate classifiers~\cite{rokach2010pattern}. Methods that use multiple base classifiers to make one decision are known as Multiple Classifier System (MCS) or Ensemble of Classifiers (EoC) \cite{zhou2012ensemble} and are one of the major development directions in machine learning. MCS were also used in many practical aspects~\cite{hang2020ensemble,li2019high,xiao2019svm} where EoC proved to have a significant impact on the performance of recognition systems. In general, the procedure for creating an EoC can be divided into three major steps~\cite{kuncheva2014combining}:

    \begin{itemize}
        \item Generation -- a phase where individual classifiers are trained~\cite{santucci2017parameter}.
        \item Selection -- a phase where only several (or one) individual models from the previous step are selected to EoC~\cite{cruz2018dynamic}.
        \item Combining -- a process of combining outputs of base classifiers to obtain an integrated model of classification~\cite{mohandes2018classifiers}.
    \end{itemize}

The combining phase can be achieved using different types of the classifier's output: a class label~\cite{kuncheva2014combining}, a subset of labels ordered by plausibility~\cite{przybyla2017dispersed}, a vector of all possible labels with corresponding scoring functions~\cite{wozniak2013hybrid}. The scoring function may have a different nature depending on the type of individual learners. The generative classification models are probabilistic in nature, and therefore return joint probability distribution over the examples and the class labels directly. The discriminative base learners, on the other hand, focus only on the conditional relation of a class label to the given example. In this case of individual learners, non-probabilistic scores are used as classifier output. Because the various outputs have to be made comparable beforehand to represent the scoring functions in a common space, the calibration method is applied~\cite{naeini2018binary,xu2016evidential}. The approaches of classifier output calibration focus on: continuousness, non-decreasingness, universal flexibility, and computational tractability~\cite{wang2019calibrating}.

The score function proposed in~\cite{NIPS2018_7798} provides information about the relative position of the recognized object in the feature space. This method uses the nearest neighborhood of the recognized object to calculate the score function value.

In this work, we propose a novel approach for calculating a scoring function based on the distance of the recognized object from the decision boundary of a given base classifier and the distance to the class centroid. Accordingly, we proposed that the new method for determining the scoring function takes into consideration not only the classifier's geometrical properties of the decision boundary created by the base learner but also the geometrical properties of the input space (training subset). Therefore, our approach provides information about the relative position of the recognized object in the feature space, which depends on class centroids and not on the nearest neighborhood of the recognized object.

Given the above, the main objectives of this work can be summarized as follows:
    \begin{itemize}
        \item A proposal of a new scoring function that uses the location of the class label centroid and the distance to the decision boundary determined by the linear classifier.
        \item The proposed score function has the same nature for all linear base classifiers, which means that outputs of these classifiers are equally represented and have the same meaning.
        \item An experimental setup to compare the proposed method with an approach that uses only the distance to the decision boundary of the base classifier to calculate the value of a score function, as well as with other existing combining base classifiers methods using different performance measures.
    \end{itemize}

The outline of the paper is as follows: In the next section (Section~\ref{sec:RelWork}), related works are presented. The proposed algorithm is presented in Section~\ref{sec:ProposedMethod}. In Section \ref{sec:ExpSetup}, the experiments that were carried out are presented, while results and discussion are present in Section~\ref{sec:Results}. Finally, we conclude the paper and propose some future works in Section~\ref{sec:Conclusions}.

\section{Related Work}\label{sec:RelWork}

In general, a classifier is a function that maps the feature space $\mathbb{X}$ into a set of class labels $\mathbb{M}$~\cite{duda2012pattern}. Usually, if we talk about a classifier, we implicitly assume that the classifier is built using some kind of supervised learning procedure. That is a procedure that incorporates information extracted from the training set~\cite{duda2012pattern}. The training set consists of training samples (taken from the feature space) and information about the class points to which these samples belong to. There are many types of classifiers such as statistical classifiers~\cite{Devroye1996}, classification trees~\cite{Quinlan1993}, SVM-based classifiers~\cite{Cortes1995} and neural networks~\cite{Zhao2019a} to name only a few. 

The choice of feature space is also important. A proper choice of the feature space may significantly improve the classification results and the wrong choice may made the classification problem harder~\cite{duda2012pattern, guyon2008feature}.  The process of choosing or creating a proper set of features is called feature extraction~\cite{guyon2008feature}.  The literature presents many techniques for attribute extraction such as: PCA~\cite{jolliffe2016principal}, kernel PCA~\cite{Schlkopf1998}, isomap~\cite{balasubramanian2002isomap}, or deep neural networks~\cite{Zhao2019a,Zhao2020,Zhao2019b}.

In this paper, it is assumed that the input space $\mathbb{X}$ is a $d-\mathrm{dimensional}$ Euclidean space $\mathbb{X}=\mathbb{R}^d$. Each object from the input space $x\in\mathbb{X}$ belongs to one of two available classes, so the output space is: $\mathbb{M}=\left\{-1;1\right\}$. It is assumed that there exists an unknown mapping $f:\mathbb{X}\mapsto\mathbb{M}$ that assigns all input space coordinates into a proper class. A classifier $\psi:\mathbb{X}\mapsto\mathbb{M}$ is a function that is designed to provide an approximation of the unknown mapping $f$. 

\subsection{Linear Binary Classifier}

A linear classifier separates the classes using a hyperplane $\pi$ defined by the following equation:
\begin{align}
 \pi: \dotProd{n}{x} + b &=0,
\end{align}
where $n$ is a unit normal vector of the decision hyperplane ($\vnorm{n}=1$), $b$ is the distance from the hyperplane to the origin and $\dotProd{\cdot}{\cdot}$ is a dot product defined as follows~\cite{Kostrikin2005}:
\begin{align}
 \dotProd{a}{b} = \sum_{i=1}^{d}a_{i}b_{i},\;\forall a,b \in \mathbb{X}.
\end{align}
If not stated otherwise, a norm of the vector $x$ is defined using the dot product:
\begin{align}\label{eq:dotP}
 \vnorm{x} &= \sqrt{\dotProd{x}{x}}.
\end{align}

The linear classifier makes its decision according to the following rule:
\begin{align}\label{eq:LinClass}
 \psi(x) &= \sign \left( \omega(x) \right),
\end{align}
where $\omega(x) = \dotProd{n}{x} + b$ is the so called \textit{discriminant function} of the classifier $\psi$~\cite{kuncheva2014combining}. When the normal vector of the plane is a unit vector, the absolute value of the discriminant function equals the perpendicular distance from the decision hyperplane to the point $x$ (the shortest distance from the plane to $x$). The sign of the discriminant function depends on the site of the plane where the instance $x$ lies. 

The decision-plane coefficients are obtained in a~supervised learning procedure (training procedures used during the experimental evaluation are mentioned in Section~\ref{sec:ExpSetup}) using the training set $\mathcal{T}$ containing $|\mathcal{T}|$ pairs of feature vectors ${x}$ and corresponding labels ${m}$:
\begin{equation}\label{eq:trainSet} 
\mathcal{T}=\left\{({x}^{(1)},m^{(1)}), ({x}^{(2)},m^{(2)}), \ldots ,({x}^{(|\mathcal{T}|)},m^{(|\mathcal{T}|)})\right\},
\end{equation}
where ${x}^{(k)} \in \mathbb{X}$ and $m^{(k)} \in \mathbb{M}$.

\subsection{Ensemble of classifiers}

Now, let us determine an ensemble of classifiers:
\begin{align}
  {\Psi} &=\setBracks{\psi^{(1)},\psi^{(2)},\cdots,\psi^{(N)}}
\end{align}
that is a set of $N$ classifiers that work together in order to produce a more robust result~\cite{kuncheva2014combining}. There are multiple strategies to combine the classifiers constituting the ensemble of linear classifiers. The simplest strategy to combine the outcomes of multiple classifiers is to apply the majority voting scheme:
\begin{align}
 \label{eq:crispVoting}\omega(x) &= \sum_{i=1}^{N}\sign(\omega^{(i)}(x)),
\end{align}
where $\omega^{(i)}(x)$ is the value of the discriminant function provided by the classifier $\psi^{(i)}$ for point $x$. However, this simple yet effective strategy completely ignores the distances of the instance $x$ from the decision planes.

Another strategy is model averaging~\cite{Skurichina1998}. The output of the averaged model may be calculated by simply averaging the values of the discriminant functions:
\begin{align}
 \label{eq:softVoting} \omega(x) &= \frac{1}{N}\sum_{i=1}^{N}\omega^{(i)}(x)
\end{align}

An alternative strategy is to normalize the discriminant function within the interval $\left[0,1\right]$. The normalization may be done using a sigmoid function --  the \textit{softmax} function for example~\cite{kuncheva2014combining}:
\begin{align}
 \label{eq:softMax} \widetilde{\omega}^{(i)}(x) &= \left( 1+ \exp (-\omega^{(i)}(x)) \right)^{-1}.
\end{align}
 The value of the discriminant function may also be used to estimate the conditional probability of a class given the instance $x$~\cite{platt1999probabilistic,Zadrozny2002}. The normalized outputs are then simply averaged:
\begin{align}
 \label{eq:softVoting2} \omega(x) &=\frac{1}{N}\sum_{i=1}^{N}  \widetilde{\omega}^{(i)}(x).
\end{align}

After combining the base classifiers, the final prediction of the ensemble is obtained according to the rule~\eqref{eq:LinClass}.

\subsection{Distance-based approaches to supervised classification}

The distance-based approaches are often used to determine the scoring functions or other functions in supervised classification. In linear Support Vector Machines algorithms, the distance to the separating decision boundary is used to compute the scoring function. Afterwards, the calibration converts scores functions into a probability measure, or more precisely transforms classifier outputs into values that can be interpreted as probabilities. The calibration methods can be generally divided into two groups: parametric and non-parametric methods. The sigmoidal transformation maps the score function to a calibrated probability output as was proposed by Platt~\cite{platt1999probabilistic}. This type of calibration assumes that the distances on either side of the decision boundary are normally distributed. The non-parametric methods are based on binning~\cite{Zadrozny2001} or isotonic regression ~\cite{Zadrozny2002}.

The nearest neighbor methods also use a distance-based approach, which is closely related to the nature of the nearest distance~\cite{ekin1999distance}. In general, these methods classify the recognized object to the class label for which the sum of the distances from its reference group of objects is the smallest. The decision boundary is determined by these discriminatory methods. The points in the feature space defining the decision boundary are characterized by the same distance from the reference group of objects for each class label. Thus, the nearest neighbor methods do not take into account the distance from the decision boundary.

In this paper, the distance-based approaches are used to determine the trust (scoring functions) of the base of the classifier~\cite{NIPS2018_7798}. The proposed algorithm calculates the ratio between the distance from the recognized object to the nearest class different from the predicted class label and the distance to the predicted class label. This approach also does not take into account the distance from the decision boundary, but only the distance from objects located near the recognized object.

The impact of distance from the decision boundary is also associated with a classification error in boosting algorithms. In particular, hybrid weighting methods that modulate the emphasis on objects according to their distance to the decision boundary have been proposed. The work~\cite{gomez2010committees} proposes an emphasis function in which the first term takes large values for patterns with large quadratic error, and the second term increases for objects that lie close to the decision boundary. There has also been a proposal~\cite{ahachad2017boosting} that the emphasis function balances also the contribution of the error and the distance to the decision boundary.

\section{Proposed Method}\label{sec:ProposedMethod}



First of all, it must be understood that the model-averaging approach has a major drawback. That is, the discriminant function of a linear classifier is unbounded. Consequently, a decision plane placed far from the real decision boundary will produce a high value of the discriminant function that may negatively affect the ensemble. For the same reason, the outliers may acquire an abnormally high value of the discriminant function which may also affect the decision of the ensemble. We can get rid of this disadvantage by applying the majority-voting scheme. However, this combination scheme loses some information by ignoring the distance to the decision plane. A compromise between the above-mentioned methods is to transform the discriminant function by applying a kind of sigmoid (the soft-max function, for example~\cite{kuncheva2014combining}) function to it. The sigmoid function is a monotonic function that has finite upper and lower bounds. As a consequence, the distance-specific information is not lost, and the impact of the misplaced decision boundaries is reduced. What is more, this approach is more flexible since the steepness of the sigmoid function may be controlled.

\begin{figure}[tb]
 \centering
  \includegraphics[width = 0.8\textwidth]{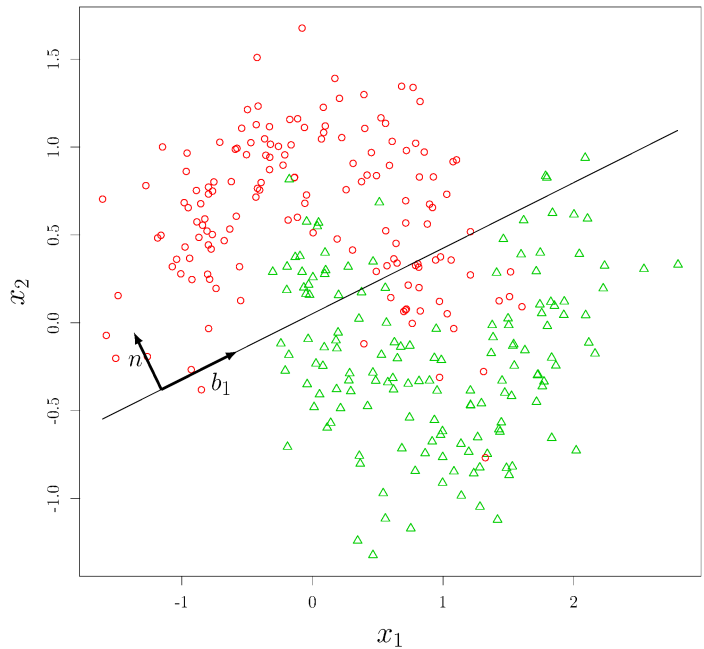}
  \caption{Linear decision boundary created by the Nearest Centroid classifier for binary, two-dimensional, banana-shaped data. Objects belonging to the first class have been marked using red circles, whereas points belonging to the other class have been marked using green triangles. The plot also shows the normal vector of the decision plane $n$ and a base vector of the plane $b_1$.  \label{fig:LinClassSim}}
\end{figure}

The other issue with combining linear classifiers is that the discriminant function of the linear classifier grows monotonically with the distance to the decision plane. It means that the linear classifier ignores the data spread along the normal vector of the decision plane and it implicitly assumes that the distribution is uniform. However, in many real-world datasets objects are distributed in various areas, and there are no objects outside these areas. An example of this situation is visualised in \figurename~\ref{fig:LinClassSim}.  The figure presents a binary, two-dimensional, banana-shaped dataset and the decision boundary created using the Nearest Centroid classifier. As we can see, the objects are placed in one cluster located in the intersection of intervals $x_1 \in [-1.5;2.5],\ x_2 \in [-1.5;2]$. Outside this area, there are no class-specific objects. Consequently, the discriminant function generated by the classifier should be low outside this area. Unfortunately, a linear classifier ignores this fact and its support will grow (along the normal vector $n$ of the decision boundary) outside this area. Transforming the discriminant function using a monotonic function, such as sigmoid function, does not change the situation at all. This is because far from the decision boundary the discriminant function approaches its upper (lower) limit. Being close to the limit still indicates high support for a particular class in the area where there are no class-specific instances. 

\begin{figure}[tb]
 \centering
  \includegraphics[width = 0.7\textwidth]{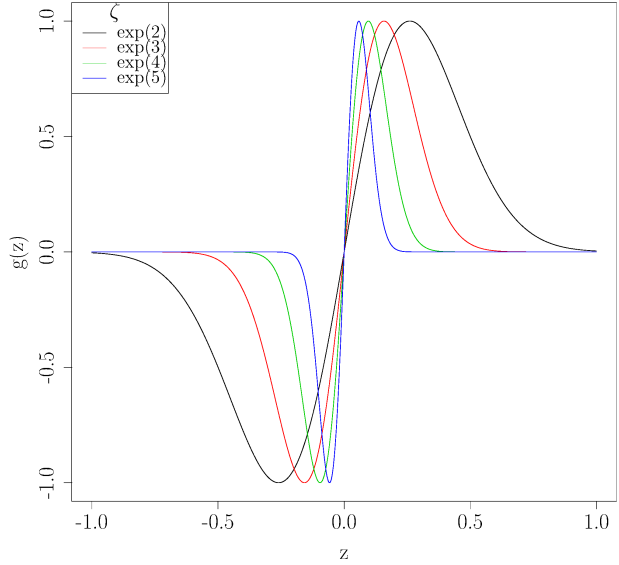}
  \caption{Visualisation of the potential function $g(z)$ (see~\eqref{eq:Potential2}) for different values of $\neighCoeff$.  The values along z-axis represent values of the discriminative function of a linear classifier. The values along the $g(z)-\mathrm{axis}$ represent the transformed values.  \label{fig:G2PotFunction}}
\end{figure}

Ignoring the class-spread-related information does not change the outcome of the single classifier since the sign of the discriminant function remains the same. However, our previous research has shown that employing this information may improve the classification quality for heterogeneous ensembles of linear classifiers~\cite{Trajdos2019}. In the previously-proposed approach, the discriminant function is transformed using a non-monotonic function derived below:  
\begin{align}\label{eq:Potential2}
 g(z) &= z\exp(-\neighCoeff z^{2} +0.5)\sqrt{2\neighCoeff},
\end{align}
where $\neighCoeff$ is a coefficient that determines the position and steepness of peaks (positive and negative peaks) (see \figurename~\ref{fig:G2PotFunction}). This coefficient should be tuned during the training procedure. The translation constants $0.5$ and the scaling factor $\sqrt{2\neighCoeff}$ guarantee that the maximum and minimum values (peaks) of the discriminant functions are $1$ and $-1$ respectively. The above-mentioned non-monotonic function is visualised in \figurename~\ref{fig:G2PotFunction}. The figure shows the shape of the function for different values of $\neighCoeff$. As we can see in the figure:
\begin{itemize}
 \item The sign of the function value is the same as the sign of its argument. As a consequence, the transformation does not change the prediction of a single classifier.
 \item The function is an odd function, meaning $g(-z) = -g(z)$.
 \item For arguments close to zero, the function also achieves values near zero. 
 \item When the function argument increases, the function value also increases until it reaches a peak.  Then the function value decreases and it approaches zero. 
\end{itemize}

Using this transformation, the prediction of the ensemble is calculated as follows:
\begin{align}\label{eq:transResponse}
 \omega(x) & = \sum_{i=1}^{N}g(\omega^{(i)}(x)).
\end{align}
Harnessing the above-mentioned transformation allows the ensemble to improve the classification quality. This is due to the function being tuned so that the potential is near zero in the areas where there are no training points. However, when the data distribution is imbalanced, the performance may degrade~\cite{Trajdos2019}. The other drawback of this method is that the $\neighCoeff$ coefficient controls the position and the steepness of the peaks simultaneously. This can be seen in \figurename~\ref{fig:G2PotFunction}. The higher the value of $\neighCoeff$ is, the peak is closer to the decision boundary, and it is narrower.  The solution may be to use an asymmetric, data-driven transformation. This transformation should place peaks of the function over class centroids. 

The linear classifiers also ignore the information about data distribution along the vectors of the plane basis since they use only information about the distance between an object and the plane. The basis is a set of linearly independent vectors that span the plane~\cite{Kostrikin2005}:
\begin{align}
 B &= \setBracks{b_1,b_2,\cdots, b_{d-1}}.
\end{align}
\figurename~\ref{fig:LinClassSim} shows a base vector for the decision boundary for two-dimensional data. As we can see in the figure, the data distribution along the base vector is not uniform in the entire feature space. Unfortunately, a linear classifier is unable to use this information.

In this work, we propose a combination method that uses information about the data spread along the normal vector of the decision plane and the spread along the plane basis as well. To do so, we use information about the data spread around the class-specific centroid:
\begin{align}\label{eq:ClassCentroid}
 C_{m} &= \frac{1}{|\mathcal{T}_{m}|}\sum_{x \in \mathcal{T}_{m}}x,
\end{align}
where $\mathcal{T}_{m}$ is a set of points belonging to class $m$:
\begin{align}\label{eq:TrainingSubset}
 \mathcal{T}_{m} &= \setBracks{ x^{(k)} | x^{(k)} \in \mathcal{T} \wedge m^{(k)} = m }
\end{align}

The distance from the class centroid $C_m$ to the instance $x$ is derived as the class-specific Mahalanobis~\cite{Zhao2015} distance:
\begin{align}\label{eq:CentroidDistance}
 \classMahDist_{m}(x) &= \sqrt{(x-C_m)^{T}\classCovar_{m}^{-1}(x-C_m)},
\end{align}
where $S_m$ is the covariance matrix of the cloud of points belonging to $\mathcal{T}_m$. The Mahalanobis distance is used because it is reported to be more efficient in finding the well-separated clusters of points in the feature space~\cite{Zhao2015}. This is an important property, because such a measure allow us to find a better description of a class-related cluster of points. 

\begin{figure}[tb]
    \centering
  {\includegraphics[width = 0.6\textwidth]{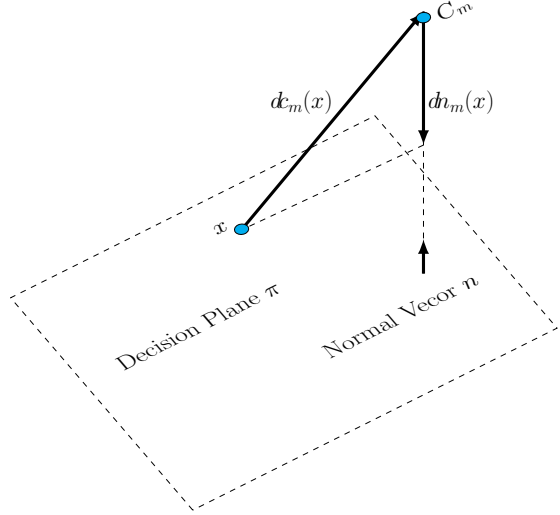}}
  \caption{Visualization of distances from an instance $x$ to the class centroid $C_m$ used to construct the proposed classifier. $\pi$ is the decision plane generated by a linear classifier. The normal vector of the plane is denoted as $n$. $\classMahDist_{m}(x)$ is the Mahalanobis distance between $x$ and $C_m$. $\classNormDist_{m}(x)$ is the length (in terms of the Mahalanobis distance) of the vector $x-C_m$ projected onto $n$ \label{fig:expDiag}}
\end{figure}

Additionally, the distance that uses information about the data spread along the normal vector is also used: 
\begin{align}
 \classNormDist_{m}(x) &= \sqrt{\projection{(x-C_m)}{n}^{T}\classProjCovar_{m}^{-1}\projection{(x-C_m)}{n}}, 
\end{align}
where $\classProjCovar_m$ is the covariance matrix of the cloud of class-specific points projected onto a normal vector $n$. The projection onto the normal vector is defined as follows:
\begin{align}\label{eq:projection}
\projection{x}{n} &= \frac{\dotProd{x}{n}}{\vnorm{n}}x.
\end{align}
The above-mentioned distances are visualized in the \figurename~\ref{fig:expDiag}.

\begin{figure}[tb]
  \centering
  {\includegraphics[width = 0.7\textwidth]{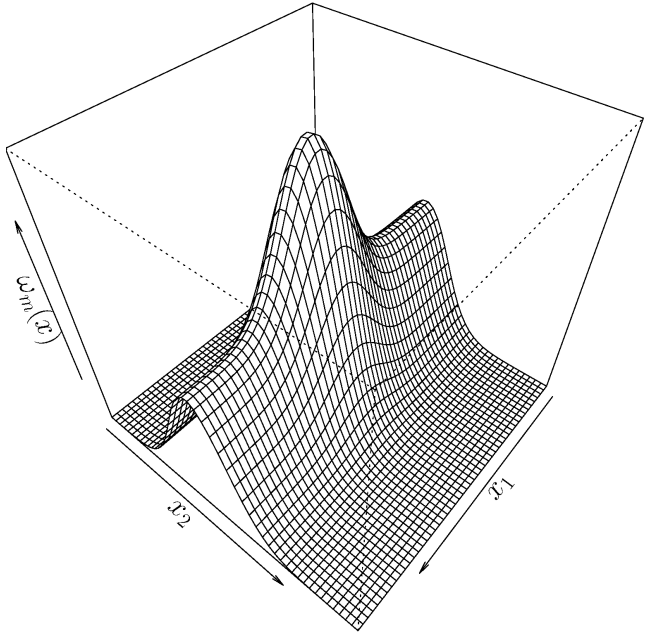}}
  \caption{Example plot of the class-specific potential function $\omega_{m}(x)$. Axes $x_1$ and $x_2$ span two dimensional space $\mathbb{X}=\mathbb{R}^2$. The values of the potential function are presented using axes denoted as $\omega_{m}(x)$.  Parameters of the potential function are set as follows: $\beta=0.5, \gamma=0.25$. The decision limit is a straight line with equation $x_1=0$. The ridge parallel to the $x_1$ axis is related to the distance between $x$ and the decision boundary $\classNormDist_{m}(x)$. The peak is related to distance between $x$ and class centroid $\classMahDist_{m}(x)$. \label{fig:PotentialExample}}
\end{figure}

The class-specific potential is then calculated using the following formula:
\begin{align}\label{eq:ProposedRule}
 \omega_{m}(x) &=   \beta \exp(-\gamma \classMahDist_{m}(x)^2) + (1-\beta)\exp(-\gamma \classNormDist_{m}(x)^{2}),
\end{align}
where $\beta \in \left[ 0,1 \right]$ is the proportion in which potentials arise from $\classMahDist_{m}(x)$ and  $\classNormDist_{m}(x)$ are mixed. 
The other coefficient $\gamma \in \mathbb{R}$ is responsible for the steepness of the peak and ridge (see \figurename~\ref{fig:PotentialExample}).
Those coefficients must be tuned for each dataset. An example of the class-specific potential function generated for the linear classifier in the two-dimensional space is shown in \figurename~\ref{fig:PotentialExample}.
Then, the potential value (the value of the discriminant function) for the point $x$ is calculated:
\begin{align}\label{eq:PropPotentialFull}
 \omega(x) &= \sum_{m \in \mathbb{M}}m\omega_{m}(x).
\end{align}
The single-classifier prediction is made using the formula~\eqref{eq:LinClass}. If the classifier works in an ensemble, the ensemble-specific value of the discriminant function is obtained using the model-averaging approach~\eqref{eq:softVoting}.

\section{Experimental Setup}\label{sec:ExpSetup}
This section provides a detailed description of the experimental setup. 

\subsection{Design of the Experiment}\label{sec:ExpSetup:ExpSetup}

In the experimental study we conducted, the proposed approach was used to combine classifiers using a homogeneous ensemble of classifiers. The ensembles were created using a bagging approach~\cite{Skurichina1998}. The generated ensembles consist of 11 classifiers learned by using the bagging method containing $80\%$ of the number of instances from the original dataset. 

During the experiment, the following ensembles were considered:
\begin{itemize}
 \item $\psi_{\mathrm{SM}}$ -- outputs of the linear classifiers were normalized using the softmax rule, and then combined according to~\eqref{eq:softVoting2}.
 \item $\psi_{\mathrm{MA}}$ -- bagged classifiers were combined using the model-averaging approach~\eqref{eq:softVoting}.
 \item $\psi_{\mathrm{MV}}$ -- bagged classifiers were combined using majority-voting approach~\eqref{eq:crispVoting}.
 \item $\psi_{\mathrm{PF}}$ -- the approach using the potential function proposed in~\cite{Trajdos2019}. Classifiers were combined using model averaging.
 \item $\psi_{\mathrm{PC}}$ -- the proposed approach.
\end{itemize}

The following base classifiers were used to build the above-mentioned ensembles:
\begin{itemize}
 \item $\psi_{\mathrm{FLDA}}$ -- Fisher~LDA~\cite{McLachlan1992},
 \item $\psi_{\mathrm{LR}}$ -- Logistic regression classifier~\cite{Devroye1996},  
 \item $\psi_{\mathrm{MLP}}$ -- single layer MLP classifier~\cite{gurney1997an},
 \item $\psi_{\mathrm{NC}}$ -- nearest centroid (Nearest Prototype)~\cite{Kuncheva1998} with the class-specific Mahalanobis distance,
 \item $\psi_{\mathrm{SVM}}$ -- SVM classifier with linear kernel (no kernel)~\cite{Cortes1995}.
\end{itemize}

The classifiers used were implemented in the WEKA framework~\cite{Hall2009}. If not stated otherwise, the classifier parameters were set to their defaults. The multi-class problems were dealt with using One-vs-One decomposition~\cite{Hllermeier2010}. The source code of the proposed algorithms is available online~\footnote{\url{https://github.com/ptrajdos/piecewiseLinearClassifiers/tree/master}}.

The $\beta$ and $\gamma$ coefficients (see equation~\eqref{eq:ProposedRule}) are tuned using the grid search approach in such a way that it maximizes the kappa coefficient~\cite{Banerjee1999}. The following sets of values are considered $\beta \in \setBracks{0,0.1,0.2,\cdots,1.0}$, $\gamma \in \setBracks{2^{-2}, 2^{-1}, \cdots , {2^{2}}}$.

To evaluate the proposed methods, six classification-quality criteria are used. The criteria are described in section~\ref{sec:ExpSetup:qualityMeasures}.

Following the recommendation of~\cite{garcia2008extension} the statistical significance of the obtained results was assessed using the two-step procedure. The first step was to perform the Friedman test~\cite{garcia2008extension} for each quality criterion separately. Since the multiple criteria were employed, the family-wise errors (\FWER{}) should be controlled~\cite{Bergmann1988}. To do so, the Bergmann-Hommel~\cite{Bergmann1988} procedure of controlling \FWER{} of the conducted Friedman tests was employed. When the Friedman test shows that there is a significant difference within the group of classifiers, the pairwise tests using the Wilcoxon signed-rank test~\cite{garcia2008extension} were employed. To control \FWER{} of the Wilcoxon-testing procedure, the Bergmann-Hommel approach was employed~\cite{Bergmann1988}. For all tests, the significance level was set to $\alpha=0.05$.

\tablename~\ref{tab:BenchmarkSetsCharacteristics} displays the collection of the $43$ benchmark sets that were used during the experimental evaluation of the proposed algorithms. The table is divided into three columns.  Each column is organized as follows. The first column contains the names of the datasets. The remaining ones contain the set-specific characteristics of the benchmark sets: the number of instances in the dataset ($|S|$); dimensionality of the input space ($d$); the number of classes ($C$); average imbalance ratio ($\mathrm{IR}$).
{
\def\arraystretch{1.3}%
\begin{sidewaystable}
 \centering\scriptsize
 \caption{The characteristics of the benchmark sets}\label{tab:BenchmarkSetsCharacteristics}
 \begin{tabular}{lcccc|lcccc|lcccc}
Name&$|S|$&$d$&$C$&$\mathrm{IR}$&Name&$|S|$&$d$&$C$&$\mathrm{IR}$&Name&$|S|$&$d$&$C$&$\mathrm{IR}$\\
\hline
appendicitis&106&7&2&2.52&housevotes&435&16&2&1.29&shuttle&57999&9&7&1326.03\\
australian&690&14&2&1.12&ionosphere&351&34&2&1.39&sonar&208&60&2&1.07\\
balance&625&4&3&2.63&iris&150&4&3&1.00&spambase&4597&57&2&1.27\\
banana2D&2000&2&2&1.00&led7digit&500&7&10&1.16&spectfheart&267&44&2&2.43\\
bands&539&19&2&1.19&lin1&1000&2&2&1.01&spirals1&2000&2&2&1.00\\
Breast Tissue&105&9&6&1.29&lin2&1000&2&2&1.83&spirals2&2000&2&2&1.00\\
check2D&800&2&2&1.00&lin3&1000&2&2&2.26&spirals3&2000&2&2&1.00\\
cleveland&303&13&5&5.17&magic&19020&10&2&1.42&texture&5500&40&11&1.00\\
coil2000&9822&85&2&8.38&mfdig fac&2000&216&10&1.00&thyroid&7200&21&3&19.76\\
dermatology&366&34&6&2.41&movement libras&360&90&15&1.00&titanic&2201&3&2&1.55\\
diabetes&768&8&2&1.43&newthyroid&215&5&3&3.43&twonorm&7400&20&2&1.00\\
Faults&1940&27&7&4.83&optdigits&5620&62&10&1.02&ULC&675&146&9&2.17\\
gauss2DV&800&2&2&1.00&page-blocks&5472&10&5&58.12&vehicle&846&18&4&1.03\\
gauss2D&4000&2&2&1.00&penbased&10992&16&10&1.04&Vertebral Column&310&6&3&1.67\\
gaussSand2&600&2&2&1.50&phoneme&5404&5&2&1.70&wdbc&569&30&2&1.34\\
gaussSand&600&2&2&1.50&pima&767&8&2&1.44&wine&178&13&3&1.23\\
glass&214&9&6&3.91&ring2D&4000&2&2&1.00&winequality-red&1599&11&6&20.71\\
haberman&306&3&2&1.89&ring&7400&20&2&1.01&winequality-white&4898&11&7&82.94\\
halfRings1&400&2&2&1.00&saheart&462&9&2&1.44&wisconsin&699&9&2&1.45\\
halfRings2&600&2&2&1.50&satimage&6435&36&6&1.66&yeast&1484&8&10&17.08\\
hepatitis&155&19&2&2.42&Seeds&210&7&3&1.00&&&&&\\
HillVall&1212&100&2&1.01&segment&2310&19&7&1.00&&&&&\\
\end{tabular}
\end{sidewaystable}
}

The datasets come from the Keel~\footnote{\url{https://sci2s.ugr.es/keel/category.php?cat=clas}} repository or are generated by us.
The datasets are available online~\footnote{\url{https://github.com/ptrajdos/MLResults/blob/master/data/slDataFull.zip}}.

During the dataset-preprocessing stage, a few transformations on the datasets were applied. The PCA method was applied and the percentage of covered variance was set to $0.95$. The attributes were also normalized to have zero mean and unit variance. 

\subsection{Quality Measures}\label{sec:ExpSetup:qualityMeasures}

To asses the classification quality offered by the proposed method, six quality criteria are employed
\begin{itemize}
 \item Macro-averaged:
 \begin{itemize}
  \item  false discovery rate  ($1- \mathrm{precision},$$\fdr$),
  \item false negative rate ($1-\mathrm{recall}$,$\fnr$),
  \item $\fOne^{\mathrm{loss}}$ 
 \end{itemize}
 \item Micro-averaged:
 \begin{itemize}
  \item  false discovery rate  ($1- \mathrm{precision},$$\fdr$),
  \item false negative rate ($1-\mathrm{recall}$,$\fnr$),
  \item $\fOne^{\mathrm{loss}}$ 
 \end{itemize}

\end{itemize}
Macro and micro averaged measures were used to assess the performance for the majority and minority classes. This is because the macro-averaged measures are more sensitive to the performance for minority classes~\cite{Sokolova2009}. The criteria are bounded in the interval $[0,1]$, where zero denotes the best classification quality.

The exact values of the criteria can be calculated using entries of the class-specific confusion matrix created using the testing dataset $\mathcal{R}$. To create the class-specific confusion matrix, the multi-class classification problem is decomposed using One-vs-Rest technique. An example of the above-mentioned confusion matrix is shown in \tablename~\ref{table:confmatrix}.  The rows of the matrix correspond to the ground-truth classes, whereas the columns match the outcome of the classifier. The entries denoted as $\mathrm{TP}_i$, $\mathrm{TN}_i$, $\mathrm{FP}_i$, $\mathrm{FN}_i$ are class-specific true positive, true negative, false positive, and false negative rates , respectively. The above-mentioned values are calculated as follows:
\begin{align}
 \mathrm{TP}_i &= \sum_{x^{(k)}\in \mathcal{R}}\ivBrack{ \Psi(x^{(k)}) = i \wedge m^{(k)} = i)},\\
 \mathrm{TN}_i &= \sum_{x^{(k)}\in \mathcal{R}}\ivBrack{ \Psi(x^{(k)}) \neq i \wedge m^{(k)}  \neq i)},\\
 \mathrm{FP}_i &= \sum_{x^{(k)}\in \mathcal{R}}\ivBrack{ \Psi(x^{(k)}) = i \wedge m^{(k)} \neq i)},\\
 \mathrm{FN}_i &= \sum_{x^{(k)}\in \mathcal{R}}\ivBrack{ \Psi(x^{(k)}) \neq i \wedge m^{(k)} = i)},
\end{align}
where  $\ivBrack{\cdot}$ is the Iverson bracket -- function defined as:

\begin{equation}\label{eq:iversonBrack}
 \ivBrack{i=j} = 
 \left\{ \begin{array}{rcl}
 0&\textrm{if}&i \neq j\\
 1&\textrm{if}& i=j
 \end{array} \right.
\end{equation}

\begin{table}[htb]
\centering\normalsize
\caption{The confusion matrix for a~binary classification problem.\label{table:confmatrix}}
{\begin{tabular}{cc|cc}
& & \multicolumn{2}{c}{prediction}\\
& &  $\mathrm{pos}_i$ & $\mathrm{neg}_i$\\
\hline
\multirow{2}{*}{ground truth}& $\mathrm{pos}_i$& $\mathrm{TP}_i$&$\mathrm{FN}_i$\\
& $\mathrm{neg}_i$& $\mathrm{FP}_i$&$\mathrm{TN}_i$\\
\end{tabular}
}
\end{table}

The macro-averaged measures are defined as follows:
\begin{align}
 \fdr_{\mathrm{macro}} & = \frac{1}{|\mathcal{M}|}\sum_{i=1}^{|\mathcal{M}|} \frac{\mathrm{FP}_{i}}{\mathrm{TP}_{i} + \mathrm{FP}_{i}},\\
 \fnr_{\mathrm{macro}} & = \frac{1}{|\mathcal{M}|}\sum_{i=1}^{|\mathcal{M}|} \frac{\mathrm{FN}_{i}}{\mathrm{TP}_{i} + \mathrm{FN}_{i}},\\
 \fOne_{\mathrm{macro}}^{\mathrm{loss}} & = \frac{1}{|\mathcal{M}|}\sum_{i=1}^{|\mathcal{M}|} \frac{ \mathrm{FP}_{i} + \mathrm{FN}_{i}}{2\mathrm{TP}_{i} + \mathrm{FP}_{i} + \mathrm{FN}_{i}},
\end{align}
where $\mathcal{M}$ is a set containing class numbers.

The micro-averaged measures are defined as follows:
\begin{align}
 \fdr_{\mathrm{micro}} & = \frac{ \sum_{i=1}^{|\mathcal{M}|}\mathrm{FP}_{i}}{\sum_{i=1}^{|\mathcal{M}|}\mathrm{TP}_{i} + \sum_{i=1}^{|\mathcal{M}|}\mathrm{FP}_{i}},\\
 \fnr_{\mathrm{micro}} & = \frac{\sum_{i=1}^{|\mathcal{M}|} \mathrm{FN}_{i}}{\sum_{i=1}^{|\mathcal{M}|}\mathrm{TP}_{i} + \sum_{i=1}^{|\mathcal{M}|}\mathrm{FN}_{i}},\\
 \fOne_{\mathrm{micro}}^{\mathrm{loss}} & = \frac{ \sum_{i=1}^{|\mathcal{M}|}\mathrm{FP}_{i} + \sum_{i=1}^{|\mathcal{M}|}\mathrm{FN}_{i}}{2\mathrm{TP}_{i} + \sum_{i=1}^{|\mathcal{M}|}\mathrm{FP}_{i} + \sum_{i=1}^{|\mathcal{M}|}\mathrm{FN}_{i}}
\end{align}

\section{Results and Discussion}\label{sec:Results}

This section presents the results and the discussion of the outcome of the experimental study. Due to the page limit, the full results are published online~\footnote{\url{https://github.com/ptrajdos/MLResults/blob/master/Boundaries/PR_2020.zip}}

To compare multiple algorithms on multiple benchmark sets, the average ranks approach is used. In this approach, the winning algorithm achieves rank equal '1', the second achieves rank equal '2', and so on. In the case of ties, the ranks of algorithms that achieve the same results are averaged. To provide a visualization of the average ranks the radar plots are employed. In the radar plot, each of the radially arranged axes represents one quality criterion. In the plots, the data is visualized in such a way that the lowest ranks are closer to the centre of the graph. Consequently, higher ranks are placed near the outer ring of the graph. Graphs are also scaled so that the inner ring represents the lowest rank recorded for the analyzed set of classifiers, and the outer ring is equal to the highest recorded rank. 

The numerical results are given in \tablename~\ref{table:ensembleFLDA}~to~ \ref{table:ensembleSMO}. Each table is structured as follows. The first row contains the names of the investigated algorithms. Then, the table is divided into six sections -- one section is related to a single evaluation criterion. The first row of each section is the name of the quality criterion investigated in the section. The second row shows the p-value of the Friedman test. The third one shows the average ranks achieved by algorithms. The following rows show p-values resulting from the pairwise Wilcoxon test. The p-value equal to $.000$ informs that the p-values are lower than $10^{-3}$ and p-value equal to $1.00$ informs that the value is higher than $0.999$. P-values lower than $\alpha$ are bolded.

Now let us analyze the classification quality of the previously proposed method $\psi_{\mathrm{PF}}$. In general, the results showed that the method is comparable to or worse than reference methods. The method is comparable to reference methods for $\psi_{\mathrm{FLDA}}$ and $\psi_{\mathrm{NC}}$ classifiers. This is probably because the decision boundary is produced using the distance to the class centroids. Under these circumstances, the peaks of the symmetric potential function may be easily tuned to be in the correct position. The situation changes for $\psi_{\mathrm{SVM}}$ classifier. That is, the ensemble is comparable to the reference methods in terms of macro-averaged measures, but it is worse for the micro-averaged criteria. It means that the classifier prefers the minority class. This is probably because the peak position is placed near the centroid of the minority class. The wide separation margin may also play a role in the process of finding the best $\neighCoeff$ coefficient for the ensemble. That is, even if the potential function is badly tuned, the wider margin prevents some points from being misclassified. For the remaining base classifiers ($\psi_{\mathrm{LR}}$ and $\psi_{\mathrm{MLP}}$)~$\psi_{\mathrm{PF}}$ ensemble performs worse than the reference methods. This is most likely because the potential peaks cannot be set up properly with the narrow decision margins produced by those base classifiers. In other words,  the $\neighCoeff$ coefficient cannot be set properly because the peaks are too close to the decision boundary. If too many points are placed near the decision boundary, the tuned potential function is narrow and placed near to the decision boundary. Consequently, the peak is too narrow. This is because the $\neighCoeff$ coefficient controls the peak position and its width simultaneously.

The analysis of the results related to the approach proposed in this paper starts with the $\psi_{\mathrm{NC}}$ base classifier. This is because this model is equivalent to the part of the proposed model. That is, the part uses the distance $\classMahDist_{m}(x)$ is a nearest-centroid classifier. The other distance $\classNormDist_{m}(x)$ together with the normal vector of the decision plane $n$ is responsible for creating the ridge (see \figurename~\ref{fig:expDiag}). The results for this classifier are a bit inconsistent. The Friedman test shows no significant differences among the group of the analyzed classifiers; however, the post-hoc test shows significant differences for micro-averaged measures. These results suggest that the obtained differences may be on the boundary of the statistical significance. On the other hand, the average ranks for all criteria clearly show that the proposed method gets lower ranks than the other methods. The significant improvement in terms of micro-averaged criteria shows that the proposed method can improve the classification quality of the ensemble. This improvement is significant only for majority classes. The boundary produced by $\psi_{\mathrm{NC}}$ classifier gives no additional information to the model since the decision boundary produced by the distance $\classMahDist_{m}(x)$ is the same.  The improvement is only a result of incorporating the ridge-shaped potential into the ensemble-response-creating process.

The quality of the obtained results also depends on the decision boundary used to build the model (please remember that the ridge is parallel to the decision boundary fed into the model). The $\psi_{\mathrm{SM}}^{\mathrm{FLDA}}$ classifier presents similar quality to the $\psi_{\mathrm{SM}}^{\mathrm{NC}}$. The average ranks also suggests that $\psi_{\mathrm{SM}}^{\mathrm{FLDA}}$ follow the same pattern as $\psi_{\mathrm{SM}}^{\mathrm{FLDA}}$. What is more important, it may be a bit weaker than $\psi_{\mathrm{SM}}^{\mathrm{NC}}$.  Although the intuition says that the weaker classifier should be easier to outperform, the proposed approach is unable to outperform any reference classifier. This is probably because the boundary produced by the $\psi_{\mathrm{SM}}^{\mathrm{FLDA}}$ classifier does not allow the proposed approach to producing the ridge in the right direction. The same is for $\psi_{\mathrm{MLP}}$ and $\psi_{\mathrm{SVM}}$. The average-rank-pattern produced by the base classifiers is similar however, only for $\psi_{\mathrm{SVM}}$ the performance for macro-averaged $F_1$ and FNR criteria are significantly improved by the proposed approach. For $\psi_{\mathrm{MLP}}$ the improvement is significant only for macro-averaged $\fnr$ (the same is for $\psi_{\mathrm{LR}}$). It means that the recall increases at the cost of increasing the number of false positives for the minority classes. 

The situation is a bit different for $\psi_{\mathrm{LR}}$ classifier. Although the statistical tests show no significant difference, the average ranks show that applying the proposed method may decrease the classification quality measured using micro-averaged criteria. The possible explanation of this phenomenon is that the $\psi_{\mathrm{LR}}$ classifier has the lowest average ranks for micro-averaged criteria and one of the lowest ranks according to macro-averaged criteria. Consequently, any attempt to improve the recall for the minority class harms the performance of the majority class. The other explanation is that the proposed scheme is tuned for the kappa coefficient, whereas the reference methods do not directly optimize this quality criterion.dge-shaped potential into the ensemble-response-creating process.

The quality of the obtained results also depends on the decision boundary used to build the model (please remember that the ridge is parallel to the decision boundary fed into the model). The $\psi_{\mathrm{SM}}^{\mathrm{FLDA}}$ classifier presents similar quality to the $\psi_{\mathrm{SM}}^{\mathrm{NC}}$. The average ranks also suggests that $\psi_{\mathrm{SM}}^{\mathrm{FLDA}}$ follows the same pattern as $\psi_{\mathrm{SM}}^{\mathrm{FLDA}}$. What is more important, it may be a bit weaker than the $\psi_{\mathrm{SM}}^{\mathrm{NC}}$.  Although intuition says that the weaker classifier should be easier to outperform, the proposed approach is unable to outperform any reference classifier. This is probably because the boundary produced by the $\psi_{\mathrm{SM}}^{\mathrm{FLDA}}$ classifier does not allow the proposed approach to produce the ridge in the right direction. The same is for the $\psi_{\mathrm{MLP}}$ and $\psi_{\mathrm{SVM}}$. The average-rank-pattern produced by the base classifiers is similar; however, only for the $\psi_{\mathrm{SVM}}$ does the performance for macro-averaged $F_1$ and FNR criteria are significantly improved by the proposed approach. For the $\psi_{\mathrm{MLP}}$, the improvement is significant only for the macro-averaged $\fnr$ (the same is for the $\psi_{\mathrm{LR}}$). It means that the recall increases at the cost of increasing the number of false positives for the minority classes. 

The situation is a bit different for $\psi_{\mathrm{LR}}$ classifier. Although the statistical tests show no significant difference, the average ranks show that applying the proposed method may decrease the classification quality measured using micro-averaged criteria. The possible explanation of this phenomenon is that the $\psi_{\mathrm{LR}}$ classifier has the lowest average ranks for micro-averaged criteria and one of the lowest ranks according to macro-averaged criteria. Consequently, any attempt to improve the recall for the minority class harms the performance for the majority class. The other explanation is that the proposed scheme is tuned for the kappa coefficient, whereas the reference methods do not directly optimize this quality criterion.

{
\setlength\tabcolsep{1.7pt}%
\begin{table}[htb]
\centering\scriptsize
\caption{Statistical evaluation: the Wilcoxon test for the ensembles based on the FLDA classifier.\label{table:ensembleFLDA}}
\begin{tabular}{c|ccccc|ccccc|ccccc}
  & $\psi_{\mathrm{SM}}$ & $\psi_{\mathrm{MA}}$ & $\psi_{\mathrm{MV}}$ & $\psi_{\mathrm{PF}}$ & $\psi_{\mathrm{PC}}$ & $\psi_{\mathrm{SM}}$ & $\psi_{\mathrm{MA}}$ & $\psi_{\mathrm{MV}}$ & $\psi_{\mathrm{PF}}$ & $\psi_{\mathrm{PC}}$ & $\psi_{\mathrm{SM}}$ & $\psi_{\mathrm{MA}}$ & $\psi_{\mathrm{MV}}$ & $\psi_{\mathrm{PF}}$ & $\psi_{\mathrm{PC}}$ \\ 
  \hline
Nam.&\multicolumn{5}{c|}{MaFDR}&\multicolumn{5}{c|}{MaFNR}&\multicolumn{5}{c}{MaF1}\\
Frd.&\multicolumn{5}{c|}{2.999e-01}&\multicolumn{5}{c|}{2.999e-01}&\multicolumn{5}{c}{2.999e-01}\\
Rank & 2.930 & 3.140 & 2.616 & 3.360 & 2.953 & 3.035 & 3.209 & 2.744 & 3.430 & 2.581 & 3.070 & 3.186 & 2.686 & 3.360 & 2.698 \\ 
   \hline
  $\psi_{\mathrm{SM}}$ &  & 1.00 & 1.00 & .874 & 1.00 &  & .452 & .962 & .183 & .204 &  & .657 & .657 & .325 & .657 \\ 
  $\psi_{\mathrm{MA}}$ &  &  & 1.00 & .874 & 1.00 &  &  & .587 & .183 & .204 &  &  & .657 & .337 & .657 \\ 
  $\psi_{\mathrm{MV}}$ &  &  &  & .874 & 1.00 &  &  &  & .159 & .183 &  &  &  & .180 & .657 \\ 
  $\psi_{\mathrm{PF}}$ &  &  &  &  & .874 &  &  &  &  & \textbf{.039} &  &  &  &  & .069 \\ 
   \hline
Nam.&\multicolumn{5}{c|}{MiFDR}&\multicolumn{5}{c|}{MiFNR}&\multicolumn{5}{c}{MiF1}\\
Frd.&\multicolumn{5}{c|}{2.999e-01}&\multicolumn{5}{c|}{2.999e-01}&\multicolumn{5}{c}{2.999e-01}\\
Rank & 3.035 & 3.233 & 2.721 & 3.419 & 2.593 & 3.035 & 3.233 & 2.721 & 3.419 & 2.593 & 3.035 & 3.233 & 2.721 & 3.419 & 2.593 \\ 
   \hline
  $\psi_{\mathrm{SM}}$&  & .428 & .471 & .134 & .069 &  & .428 & .471 & .134 & .069 &  & .428 & .471 & .134 & .069 \\ 
  $\psi_{\mathrm{MA}}$&  &  & .428 & .134 & .069 &  &  & .428 & .134 & .069 &  &  & .428 & .134 & .069 \\ 
  $\psi_{\mathrm{MV}}$&  &  &  & .095 & .120 &  &  &  & .095 & .120 &  &  &  & .095 & .120 \\ 
  $\psi_{\mathrm{PF}}$&  &  &  &  & \textbf{.043} &  &  &  &  & \textbf{.043} &  &  &  &  & \textbf{.043} \\ 
  \end{tabular}
\end{table}
}

{
\setlength\tabcolsep{1.7pt}%
\begin{table}[htb]
\centering\scriptsize
\caption{Statistical evaluation: the Wilcoxon test for the ensembles based on $\psi_{\mathrm{LR}}$ classifier.\label{table:ensembleLogistic}}
\begin{tabular}{c|ccccc|ccccc|ccccc}
  & $\psi_{\mathrm{SM}}$ & $\psi_{\mathrm{MA}}$ & $\psi_{\mathrm{MV}}$ & $\psi_{\mathrm{PF}}$ & $\psi_{\mathrm{PC}}$ & $\psi_{\mathrm{SM}}$ & $\psi_{\mathrm{MA}}$ & $\psi_{\mathrm{MV}}$ & $\psi_{\mathrm{PF}}$ & $\psi_{\mathrm{PC}}$ & $\psi_{\mathrm{SM}}$ & $\psi_{\mathrm{MA}}$ & $\psi_{\mathrm{MV}}$ & $\psi_{\mathrm{PF}}$ & $\psi_{\mathrm{PC}}$ \\ 
  \hline
Nam.&\multicolumn{5}{c|}{MaFDR}&\multicolumn{5}{c|}{MaFNR}&\multicolumn{5}{c}{MaF1}\\
Frd.&\multicolumn{5}{c|}{6.166e-02}&\multicolumn{5}{c|}{\textbf{5.932e-06}}&\multicolumn{5}{c}{\textbf{2.201e-02}}\\
Rank & 2.686 & 2.942 & 2.930 & 3.581 & 2.860 & 2.802 & 3.093 & 2.953 & 3.965 & 2.186 & 2.733 & 2.942 & 2.837 & 3.721 & 2.767 \\ 
   \hline
  $\psi_{\mathrm{SM}}$ &  & 1.00 & 1.00 & .173 & 1.00 &  & 1.00 & 1.00 & \textbf{.001} & \textbf{.011} &  & 1.00 & 1.00 & \textbf{.013} & .443 \\ 
  $\psi_{\mathrm{MA}}$ &  &  & 1.00 & .566 & 1.00 &  &  & 1.00 & \textbf{.011} & \textbf{.011} &  &  & 1.00 & .083 & .443 \\ 
  $\psi_{\mathrm{MV}}$ &  &  &  & .173 & 1.00 &  &  &  & \textbf{.002} & \textbf{.011} &  &  &  & \textbf{.013} & .443 \\ 
  $\psi_{\mathrm{PF}}$ &  &  &  &  & .173 &  &  &  &  & \textbf{.000} &  &  &  &  & \textbf{.013} \\ 
   \hline
Nam.&\multicolumn{5}{c|}{MiFDR}&\multicolumn{5}{c|}{MiFNR}&\multicolumn{5}{c}{MiF1}\\
Frd.&\multicolumn{5}{c|}{\textbf{2.578e-05}}&\multicolumn{5}{c|}{\textbf{2.578e-05}}&\multicolumn{5}{c}{\textbf{2.578e-05}}\\
Rank & 2.500 & 2.593 & 2.605 & 3.837 & 3.465 & 2.500 & 2.593 & 2.605 & 3.837 & 3.465 & 2.500 & 2.593 & 2.605 & 3.837 & 3.465 \\ 
   \hline
  $\psi_{\mathrm{SM}}$ &  & 1.00 & 1.00 & \textbf{.001} & .399 &  & 1.00 & 1.00 & \textbf{.001} & .399 &  & 1.00 & 1.00 & \textbf{.001} & .399 \\ 
  $\psi_{\mathrm{MA}}$ &  &  & 1.00 & \textbf{.003} & .399 &  &  & 1.00 & \textbf{.003} & .399 &  &  & 1.00 & \textbf{.003} & .399 \\ 
  $\psi_{\mathrm{MV}}$ &  &  &  & \textbf{.000} & .399 &  &  &  & \textbf{.000} & .399 &  &  &  & \textbf{.000} & .399 \\ 
  $\psi_{\mathrm{PF}}$ &  &  &  &  & .492 &  &  &  &  & .492 &  &  &  &  & .492 \\ 

  \end{tabular}
\end{table}
}

{
\setlength\tabcolsep{1.7pt}%
\begin{table}[htb]
\centering\scriptsize
\caption{Statistical evaluation: the Wilcoxon test for the ensembles based on $\psi_{\mathrm{MLP}}$ classifier.\label{table:ensembleMLP}}
\begin{tabular}{c|ccccc|ccccc|ccccc}
  & $\psi_{\mathrm{SM}}$ & $\psi_{\mathrm{MA}}$ & $\psi_{\mathrm{MV}}$ & $\psi_{\mathrm{PF}}$ & $\psi_{\mathrm{PC}}$ & $\psi_{\mathrm{SM}}$ & $\psi_{\mathrm{MA}}$ & $\psi_{\mathrm{MV}}$ & $\psi_{\mathrm{PF}}$ & $\psi_{\mathrm{PC}}$ & $\psi_{\mathrm{SM}}$ & $\psi_{\mathrm{MA}}$ & $\psi_{\mathrm{MV}}$ & $\psi_{\mathrm{PF}}$ & $\psi_{\mathrm{PC}}$ \\ 
    \hline
Nam.&\multicolumn{5}{c|}{MaFDR}&\multicolumn{5}{c|}{MaFNR}&\multicolumn{5}{c}{MaF1}\\
Frd.&\multicolumn{5}{c|}{4.715e-01}&\multicolumn{5}{c|}{\textbf{4.608e-08}}&\multicolumn{5}{c}{\textbf{1.595e-03}}\\
Rank & 3.023 & 3.209 & 2.767 & 3.209 & 2.791 & 3.093 & 3.267 & 2.686 & 3.988 & 1.965 & 3.140 & 3.256 & 2.628 & 3.651 & 2.326 \\ 
  \hline
  $\psi_{\mathrm{SM}}$ &  & 1.00 & 1.00 & 1.00 & .560 &  & .226 & .070 & \textbf{.002} & \textbf{.004} &  & .228 & .228 & \textbf{.023} & .118 \\ 
  $\psi_{\mathrm{MA}}$ &  &  & .093 & 1.00 & .347 &  &  & \textbf{.024} & \textbf{.005} & \textbf{.001} &  &  & \textbf{.046} & .228 & .062 \\ 
  $\psi_{\mathrm{MV}}$ &  &  &  & 1.00 & 1.00 &  &  &  & \textbf{.000} & \textbf{.002} &  &  &  & \textbf{.006} & .140 \\ 
  $\psi_{\mathrm{PF}}$ &  &  &  &  & \textbf{.025} &  &  &  &  & \textbf{.000} &  &  &  &  & \textbf{.001} \\ 
   \hline
Nam.&\multicolumn{5}{c|}{MiFDR}&\multicolumn{5}{c|}{MiFNR}&\multicolumn{5}{c}{MiF1}\\
Frd.&\multicolumn{5}{c|}{\textbf{1.509e-02}}&\multicolumn{5}{c|}{\textbf{1.509e-02}}&\multicolumn{5}{c}{\textbf{1.509e-02}}\\
Rank & 2.860 & 2.895 & 2.453 & 3.663 & 3.128 & 2.860 & 2.895 & 2.453 & 3.663 & 3.128 & 2.860 & 2.895 & 2.453 & 3.663 & 3.128 \\ 
   \hline
  $\psi_{\mathrm{SM}}$ &  & 1.00 & .302 & \textbf{.004} & .820 &  & 1.00 & .302 & \textbf{.004} & .820 &  & 1.00 & .302 & \textbf{.004} & .820 \\ 
  $\psi_{\mathrm{MA}}$ &  &  & .302 & \textbf{.005} & .820 &  &  & .302 & \textbf{.005} & .820 &  &  & .302 & \textbf{.005} & .820 \\ 
  $\psi_{\mathrm{MV}}$ &  &  &  & \textbf{.000} & .820 &  &  &  & \textbf{.000} & .820 &  &  &  & \textbf{.000} & .820 \\ 
  $\psi_{\mathrm{PF}}$ &  &  &  &  & .302 &  &  &  &  & .302 &  &  &  &  & .302 \\ 

  \end{tabular}
\end{table}
}

{
\setlength\tabcolsep{1.7pt}%
\begin{table}[htb]
\centering\scriptsize
\caption{Statistical evaluation: the Wilcoxon test for the ensembles based on $\psi_{\mathrm{NC}}$ classifier.\label{table:ensembleNC}}
\begin{tabular}{c|ccccc|ccccc|ccccc}
  & $\psi_{\mathrm{SM}}$ & $\psi_{\mathrm{MA}}$ & $\psi_{\mathrm{MV}}$ & $\psi_{\mathrm{PF}}$ & $\psi_{\mathrm{PC}}$ & $\psi_{\mathrm{SM}}$ & $\psi_{\mathrm{MA}}$ & $\psi_{\mathrm{MV}}$ & $\psi_{\mathrm{PF}}$ & $\psi_{\mathrm{PC}}$ & $\psi_{\mathrm{SM}}$ & $\psi_{\mathrm{MA}}$ & $\psi_{\mathrm{MV}}$ & $\psi_{\mathrm{PF}}$ & $\psi_{\mathrm{PC}}$ \\ 
     \hline
Nam.&\multicolumn{5}{c|}{MaFDR}&\multicolumn{5}{c|}{MaFNR}&\multicolumn{5}{c}{MaF1}\\
Frd.&\multicolumn{5}{c|}{3.756e-01}&\multicolumn{5}{c|}{3.756e-01}&\multicolumn{5}{c}{3.756e-01}\\
Rank & 3.023 & 3.000 & 3.302 & 3.047 & 2.628 & 2.930 & 3.105 & 3.070 & 3.360 & 2.535 & 3.047 & 3.023 & 3.349 & 3.093 & 2.488 \\ 
   \hline
  $\psi_{\mathrm{SM}}$ &  & 1.00 & .743 & 1.00 & .128 &  & .620 & .133 & .297 & .201 &  & 1.00 & .241 & 1.00 & \textbf{.021} \\ 
  $\psi_{\mathrm{MA}}$ &  &  & 1.00 & 1.00 & .171 &  &  & .620 & .620 & .133 &  &  & 1.00 & 1.00 & .094 \\ 
  $\psi_{\mathrm{MV}}$ &  &  &  & 1.00 & \textbf{.032} &  &  &  & .620 & \textbf{.040} &  &  &  & 1.00 & \textbf{.003} \\ 
  $\psi_{\mathrm{PF}}$ &  &  &  &  & .128 &  &  &  &  & \textbf{.030} &  &  &  &  & .051 \\ 
   \hline
Nam.&\multicolumn{5}{c|}{MiFDR}&\multicolumn{5}{c|}{MiFNR}&\multicolumn{5}{c}{MiF1}\\
Frd.&\multicolumn{5}{c|}{1.899e-01}&\multicolumn{5}{c|}{1.899e-01}&\multicolumn{5}{c}{1.899e-01}\\
Rank & 2.930 & 3.151 & 3.233 & 3.314 & 2.372 & 2.930 & 3.151 & 3.233 & 3.314 & 2.372 & 2.930 & 3.151 & 3.233 & 3.314 & 2.372 \\ 
   \hline
  $\psi_{\mathrm{SM}}$&  & .197 & .084 & .197 & .197 &  & .197 & .084 & .197 & .197 &  & .197 & .084 & .197 & .197 \\ 
  $\psi_{\mathrm{MA}}$&  &  & .372 & .372 & \textbf{.020} &  &  & .372 & .372 & \textbf{.020} &  &  & .372 & .372 & \textbf{.020} \\ 
  $\psi_{\mathrm{MV}}$&  &  &  & .372 & \textbf{.010} &  &  &  & .372 & \textbf{.010} &  &  &  & .372 & \textbf{.010} \\ 
  $\psi_{\mathrm{PF}}$&  &  &  &  & \textbf{.020} &  &  &  &  & \textbf{.020} &  &  &  &  & \textbf{.020} \\ 
   
  \end{tabular}
\end{table}
}

{
\setlength\tabcolsep{1.7pt}%
\begin{table}[htb]
\centering\scriptsize
\caption{Statistical evaluation: the Wilcoxon test for the ensembles based on $\psi_{\mathrm{SVM}}$ classifier.\label{table:ensembleSMO}}
\begin{tabular}{c|ccccc|ccccc|ccccc}
  & $\psi_{\mathrm{SM}}$ & $\psi_{\mathrm{MA}}$ & $\psi_{\mathrm{MV}}$ & $\psi_{\mathrm{PF}}$ & $\psi_{\mathrm{PC}}$ & $\psi_{\mathrm{SM}}$ & $\psi_{\mathrm{MA}}$ & $\psi_{\mathrm{MV}}$ & $\psi_{\mathrm{PF}}$ & $\psi_{\mathrm{PC}}$ & $\psi_{\mathrm{SM}}$ & $\psi_{\mathrm{MA}}$ & $\psi_{\mathrm{MV}}$ & $\psi_{\mathrm{PF}}$ & $\psi_{\mathrm{PC}}$ \\ 
       \hline
Nam.&\multicolumn{5}{c|}{MaFDR}&\multicolumn{5}{c|}{MaFNR}&\multicolumn{5}{c}{MaF1}\\
Frd.&\multicolumn{5}{c|}{7.472e-02}&\multicolumn{5}{c|}{\textbf{5.787e-08}}&\multicolumn{5}{c}{\textbf{3.427e-05}}\\
Rank & 3.070 & 3.198 & 3.070 & 3.326 & 2.337 & 3.151 & 3.442 & 3.151 & 3.512 & 1.744 & 3.233 & 3.267 & 3.233 & 3.349 & 1.919 \\ 
   \hline
  $\psi_{\mathrm{SM}}$&  & 1.00 & 1.00 & 1.00 & .097 &  & .795 & 1.00 & .795 & \textbf{.000} &  & 1.00 & 1.00 & 1.00 & \textbf{.005} \\ 
  $\psi_{\mathrm{MA}}$&  &  & 1.00 & 1.00 & \textbf{.025} &  &  & .795 & 1.00 & \textbf{.000} &  &  & 1.00 & 1.00 & \textbf{.000} \\ 
  $\psi_{\mathrm{MV}}$&  &  &  & 1.00 & .097 &  &  &  & .795 & \textbf{.000} &  &  &  & 1.00 & \textbf{.005} \\ 
  $\psi_{\mathrm{PF}}$&  &  &  &  & \textbf{.004} &  &  &  &  & \textbf{.000} &  &  &  &  & \textbf{.000} \\ 
   \hline
Nam.&\multicolumn{5}{c|}{MiFDR}&\multicolumn{5}{c|}{MiFNR}&\multicolumn{5}{c}{MiF1}\\
Frd.&\multicolumn{5}{c|}{1.942e-01}&\multicolumn{5}{c|}{1.942e-01}&\multicolumn{5}{c}{1.942e-01}\\
Rank & 2.837 & 2.977 & 2.837 & 3.570 & 2.779 & 2.837 & 2.977 & 2.837 & 3.570 & 2.779 & 2.837 & 2.977 & 2.837 & 3.570 & 2.779 \\ 
   \hline
  $\psi_{\mathrm{SM}}$ &  & 1.00 & 1.00 & \textbf{.026} & 1.00 &  & 1.00 & 1.00 & \textbf{.026} & 1.00 &  & 1.00 & 1.00 & \textbf{.026} & 1.00 \\ 
  $\psi_{\mathrm{MA}}$ &  &  & 1.00 & .128 & 1.00 &  &  & 1.00 & .128 & 1.00 &  &  & 1.00 & .128 & 1.00 \\ 
  $\psi_{\mathrm{MV}}$ &  &  &  & \textbf{.026} & 1.00 &  &  &  & \textbf{.026} & 1.00 &  &  &  & \textbf{.026} & 1.00 \\ 
  $\psi_{\mathrm{PF}}$ &  &  &  &  & .124 &  &  &  &  & .124 &  &  &  &  & .124 \\ 
  \end{tabular}
\end{table}
}

\begin{figure}
\centering
\includegraphics[width=.7\linewidth]{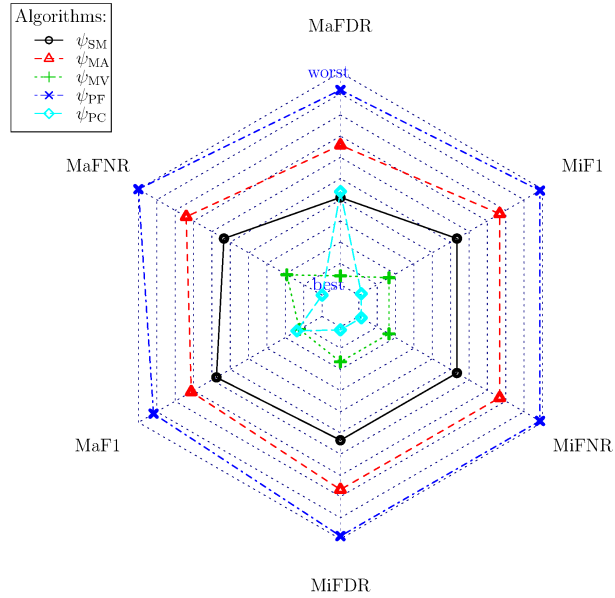}
 \caption{Radar plot for the homogeneous ensemble based on $\psi_{\mathrm{FLDA}}$ }
 \label{fig:HomoEnsembles:FLDA}
\end{figure}

\begin{figure}
\centering
\includegraphics[width=.7\linewidth]{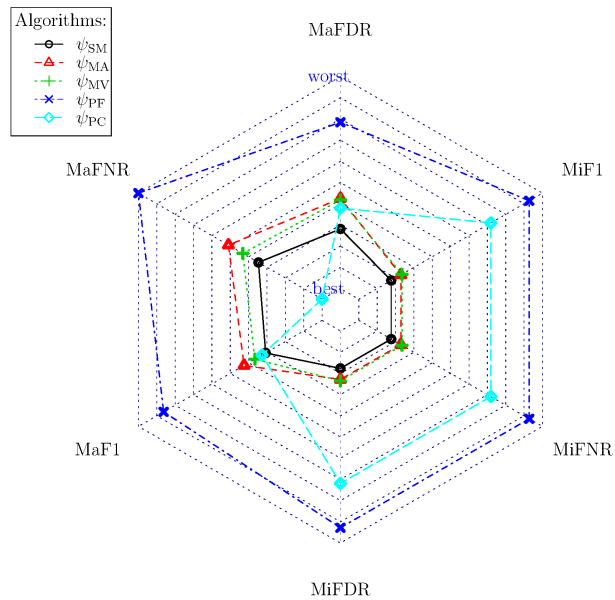}
 \caption{Radar plot for the homogeneous ensemble based on $\psi_{\mathrm{LR}}$ }
 \label{fig:HomoEnsembles:LR}
\end{figure}

\begin{figure}
\centering
\includegraphics[width=.7\linewidth]{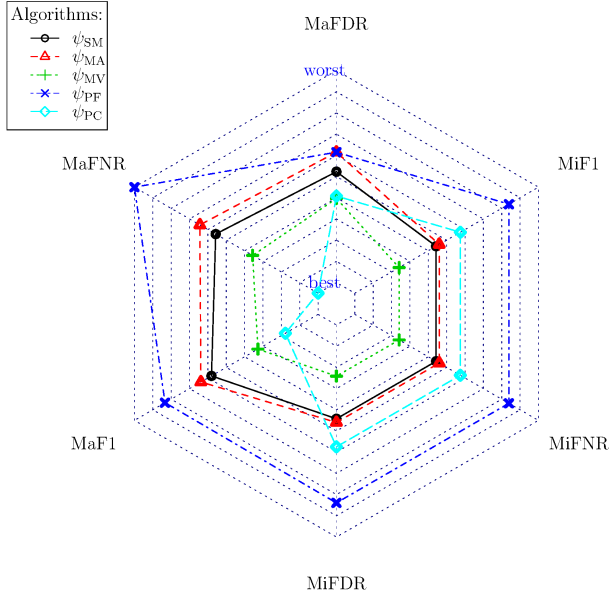}
 \caption{Radar plot for the homogeneous ensemble based on $\psi_{\mathrm{MLP}}$ }
 \label{fig:HomoEnsembles:MLP}
\end{figure}

\begin{figure}
\centering
\includegraphics[width=.7\linewidth]{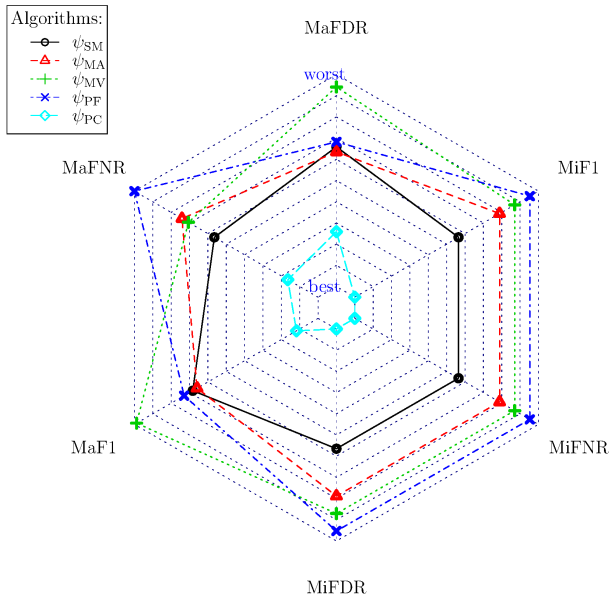}
 \caption{Radar plot for the homogeneous ensemble based on $\psi_{\mathrm{NC}}$ }
 \label{fig:HomoEnsembles:NC}
\end{figure}

\begin{figure}
\centering
\includegraphics[width=.7\linewidth]{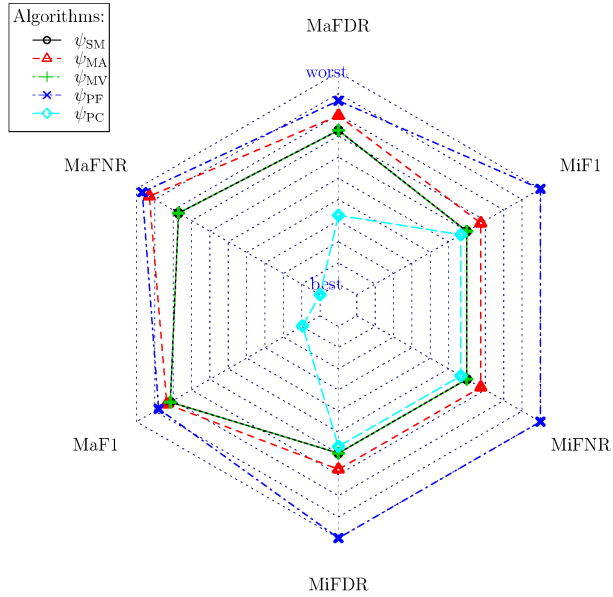}
 \caption{Radar plot for the homogeneous ensemble based on $\psi_{\mathrm{SVM}}$ }
 \label{fig:HomoEnsembles:SVM}
\end{figure}

\FloatBarrier

\section{Conclusions and Future Work}\label{sec:Conclusions}

This paper presents a new method of combining linear classifiers in the geometrical space. It means that the combination rule takes into consideration not only the classifier's outcome but also the geometrical properties of the input space and the geometrical properties of the decision boundary created by the classifiers. In the case presented, the combination rule takes into account the class-specific centroids and the normal vector of the base classifiers harnessed for the ensemble. The main purpose of employing the above-mentioned properties was to create a potential function that is spanned around the class centroids and the decision boundary. Contrary to the state-of-the-art methods, the potential function is not a monotonic function that grows with the distance to the decision boundary. The aim of using a non-monotonic function is to incorporate some information about the data distribution into the ensemble of linear classifiers. That is, the captured information is the data spread along with the decision plane and the data spread along the normal vector of the decision plane. 

The experimental results show that the proposed method may create an ensemble classifier that outperforms the commonly used methods of combining the linear classifiers. On the other hand, the proposed approach has also a few drawbacks. First of all, it uses two shape coefficients of the potential function that have to be carefully tuned. In this work, we used a grid search approach and cross-validation to find the best values of those coefficients which is rather a time-consuming approach. Consequently, our future research needs to be aimed at finding a heuristic approach that allows finding a good set of shape coefficients in a more efficient way.

\textbf{Acknowledgments.} Funding:  This work was supported by the National Science Centre, Poland under the grant no. 2017/25/B/ST6/01750.

\bibliography{bibliography}

\begin{thebibliography}{10}
\expandafter\ifx\csname url\endcsname\relax
  \def\url#1{\texttt{#1}}\fi
\expandafter\ifx\csname urlprefix\endcsname\relax\def\urlprefix{URL }\fi
\expandafter\ifx\csname href\endcsname\relax
  \def\href#1#2{#2} \def\path#1{#1}\fi

\bibitem{rokach2010pattern}
L.~Rokach, Pattern Classification Using Ensemble Methods, Vol.~75, WORLD
  SCIENTIFIC, 2009.
\newblock \href {https://doi.org/10.1142/7238} {\path{doi:10.1142/7238}}.

\bibitem{zhou2012ensemble}
Z.-H. Zhou, Ensemble Methods, Chapman and Hall/CRC, 2012.
\newblock \href {https://doi.org/10.1201/b12207} {\path{doi:10.1201/b12207}}.

\bibitem{hang2020ensemble}
J.~Hang, K.~Han, H.~Chen, Y.~Li,
  \href{https://doi.org/10.1016/j.patcog.2019.107184}{Ensemble adversarial
  black-box attacks against deep learning systems}, Pattern Recognit. 101
  (2020) 107184.
\newblock \href {https://doi.org/10.1016/j.patcog.2019.107184}
  {\path{doi:10.1016/j.patcog.2019.107184}}.
\newline\urlprefix\url{https://doi.org/10.1016/j.patcog.2019.107184}

\bibitem{li2019high}
Z.~Li, Z.~Han, J.~Xing, Q.~Ye, X.~Yu, J.~Jiao, High performance person
  re-identification via a boosting ranking ensemble, Pattern Recognition 94
  (2019) 187--195.

\bibitem{xiao2019svm}
J.~Xiao, \href{https://doi.org/10.1016/j.physa.2018.10.060}{{SVM} and {KNN}
  ensemble learning for traffic incident detection}, Physica A 517 (2019)
  29--35.
\newblock \href {https://doi.org/10.1016/j.physa.2018.10.060}
  {\path{doi:10.1016/j.physa.2018.10.060}}.
\newline\urlprefix\url{https://doi.org/10.1016/j.physa.2018.10.060}

\bibitem{kuncheva2014combining}
L.~I. Kuncheva, Combining Pattern Classifiers, John Wiley \& Sons, Inc., 2014.
\newblock \href {https://doi.org/10.1002/9781118914564}
  {\path{doi:10.1002/9781118914564}}.

\bibitem{santucci2017parameter}
E.~Santucci, L.~Didaci, G.~Fumera, F.~Roli, A parameter randomization approach
  for constructing classifier ensembles, Pattern Recognit. 69 (2017) 1--13.
\newblock \href {https://doi.org/10.1016/j.patcog.2017.03.031}
  {\path{doi:10.1016/j.patcog.2017.03.031}}.

\bibitem{cruz2018dynamic}
R.~M. Cruz, R.~Sabourin, G.~D. Cavalcanti, Dynamic classifier selection:
  {Recent} advances and perspectives, Information Fusion 41 (2018) 195--216.
\newblock \href {https://doi.org/10.1016/j.inffus.2017.09.010}
  {\path{doi:10.1016/j.inffus.2017.09.010}}.

\bibitem{mohandes2018classifiers}
M.~Mohandes, M.~Deriche, S.~O. Aliyu, Classifiers combination techniques: {A}
  comprehensive review, IEEE Access 6 (2018) 19626--19639.
\newblock \href {https://doi.org/10.1109/access.2018.2813079}
  {\path{doi:10.1109/access.2018.2813079}}.

\bibitem{przybyla2017dispersed}
M.~Przybyła-Kasperek, A.~Wakulicz-Deja, Dispersed decision-making system with
  fusion methods from the rank level and the measurement level -- a comparative
  study, Information Systems 69 (2017) 124--154.
\newblock \href {https://doi.org/10.1016/j.is.2017.05.002}
  {\path{doi:10.1016/j.is.2017.05.002}}.

\bibitem{wozniak2013hybrid}
M.~Wozniak, Hybrid Classifiers, Vol. 519, Springer Berlin Heidelberg, 2014.
\newblock \href {https://doi.org/10.1007/978-3-642-40997-4}
  {\path{doi:10.1007/978-3-642-40997-4}}.

\bibitem{naeini2018binary}
M.~Pakdaman~Naeini, G.~F. Cooper,
  \href{https://doi.org/10.1007/s10115-017-1133-2}{Binary classifier
  calibration using an ensemble of piecewise linear regression models}, Knowl
  Inf Syst 54~(1) (2017) 151--170.
\newblock \href {https://doi.org/10.1007/s10115-017-1133-2}
  {\path{doi:10.1007/s10115-017-1133-2}}.
\newline\urlprefix\url{https://doi.org/10.1007/s10115-017-1133-2}

\bibitem{xu2016evidential}
P.~Xu, F.~Davoine, H.~Zha, T.~Denœux,
  \href{https://doi.org/10.1016/j.ijar.2015.05.002}{Evidential calibration of
  binary {SVM} classifiers}, Int. J. Approximate Reasoning 72 (2016) 55--70.
\newblock \href {https://doi.org/10.1016/j.ijar.2015.05.002}
  {\path{doi:10.1016/j.ijar.2015.05.002}}.
\newline\urlprefix\url{https://doi.org/10.1016/j.ijar.2015.05.002}

\bibitem{wang2019calibrating}
Y.~Wang, L.~Li, C.~Dang,
  \href{https://doi.org/10.1109/tpami.2019.2895794}{Calibrating classification
  probabilities with shape-restricted polynomial regression}, IEEE Trans.
  Pattern Anal. Mach. Intell. 41~(8) (2019) 1813--1827.
\newblock \href {https://doi.org/10.1109/tpami.2019.2895794}
  {\path{doi:10.1109/tpami.2019.2895794}}.
\newline\urlprefix\url{https://doi.org/10.1109/tpami.2019.2895794}

\bibitem{NIPS2018_7798}
H.~Jiang, B.~Kim, M.~Guan, M.~Gupta,
  \href{http://papers.nips.cc/paper/7798-to-trust-or-not-to-trust-a-classifier.pdf}{To
  trust or not to trust a classifier}, in: S.~Bengio, H.~Wallach,
  H.~Larochelle, K.~Grauman, N.~Cesa-Bianchi, R.~Garnett (Eds.), Advances in
  Neural Information Processing Systems 31, Curran Associates, Inc., 2018, pp.
  5541--5552.
\newline\urlprefix\url{http://papers.nips.cc/paper/7798-to-trust-or-not-to-trust-a-classifier.pdf}

\bibitem{duda2012pattern}
R.~O. Duda, P.~E. Hart, D.~G. Stork, Pattern classification, John Wiley \&
  Sons, 2012.

\bibitem{Devroye1996}
L.~Devroye, L.~Györfi, G.~Lugosi, A Probabilistic Theory of Pattern
  Recognition, Springer New York, 1996.
\newblock \href {https://doi.org/10.1007/978-1-4612-0711-5}
  {\path{doi:10.1007/978-1-4612-0711-5}}.

\bibitem{Quinlan1993}
J.~R. Quinlan, {{C4.5} : {Programs} for machine learning}, Morgan Kaufmann
  Publishers Inc., San Francisco, CA, USA, 1993.

\bibitem{Cortes1995}
C.~Cortes, V.~Vapnik, Support-vector networks, Mach Learn 20~(3) (1995)
  273--297.
\newblock \href {https://doi.org/10.1007/bf00994018}
  {\path{doi:10.1007/bf00994018}}.

\bibitem{Zhao2019a}
C.~Zhao, K.~Chen, Z.~Wei, Y.~Chen, D.~Miao, W.~Wang,
  \href{https://doi.org/10.1016/j.patrec.2018.04.029}{Multilevel triplet deep
  learning model for person re-identification}, Pattern Recognit. Lett. 117
  (2019) 161--168.
\newblock \href {https://doi.org/10.1016/j.patrec.2018.04.029}
  {\path{doi:10.1016/j.patrec.2018.04.029}}.
\newline\urlprefix\url{https://doi.org/10.1016/j.patrec.2018.04.029}

\bibitem{guyon2008feature}
I.~Guyon, S.~Gunn, M.~Nikravesh, L.~A. Zadeh,
  \href{https://doi.org/10.1007/978-3-540-35488-8}{Feature Extraction}, Vol.
  207, Springer Berlin Heidelberg, 2006.
\newblock \href {https://doi.org/10.1007/978-3-540-35488-8}
  {\path{doi:10.1007/978-3-540-35488-8}}.
\newline\urlprefix\url{https://doi.org/10.1007/978-3-540-35488-8}

\bibitem{jolliffe2016principal}
I.~T. Jolliffe, J.~Cadima, Principal component analysis: {A} review and recent
  developments, Phil. Trans. R. Soc. A 374~(2065) (2016) 20150202.
\newblock \href {https://doi.org/10.1098/rsta.2015.0202}
  {\path{doi:10.1098/rsta.2015.0202}}.

\bibitem{Schlkopf1998}
B.~Schölkopf, A.~Smola, K.-R. Müller,
  \href{https://doi.org/10.1162/089976698300017467}{Nonlinear component
  analysis as a kernel eigenvalue problem}, Neural Comput. 10~(5) (1998)
  1299--1319.
\newblock \href {https://doi.org/10.1162/089976698300017467}
  {\path{doi:10.1162/089976698300017467}}.
\newline\urlprefix\url{https://doi.org/10.1162/089976698300017467}

\bibitem{balasubramanian2002isomap}
M.~Balasubramanian, \href{https://doi.org/10.1126/science.295.5552.7a}{The
  isomap algorithm and topological stability}, Science 295~(5552) (2002) 7a--7.
\newblock \href {https://doi.org/10.1126/science.295.5552.7a}
  {\path{doi:10.1126/science.295.5552.7a}}.
\newline\urlprefix\url{https://doi.org/10.1126/science.295.5552.7a}

\bibitem{Zhao2020}
C.~Zhao, X.~Wang, W.~Zuo, F.~Shen, L.~Shao, D.~Miao,
  \href{https://doi.org/10.1016/j.patcog.2019.107014}{Similarity learning with
  joint transfer constraints for person re-identification}, Pattern Recognit.
  97 (2020) 107014.
\newblock \href {https://doi.org/10.1016/j.patcog.2019.107014}
  {\path{doi:10.1016/j.patcog.2019.107014}}.
\newline\urlprefix\url{https://doi.org/10.1016/j.patcog.2019.107014}

\bibitem{Zhao2019b}
C.~Zhao, K.~Chen, D.~Zang, Z.~Zhang, W.~Zuo, D.~Miao,
  \href{https://doi.org/10.1007/s11432-019-2675-3}{Uncertainty-optimized deep
  learning model for small-scale person re-identification}, Sci. China Inf.
  Sci. 62~(12) (Nov. 2019).
\newblock \href {https://doi.org/10.1007/s11432-019-2675-3}
  {\path{doi:10.1007/s11432-019-2675-3}}.
\newline\urlprefix\url{https://doi.org/10.1007/s11432-019-2675-3}

\bibitem{Kostrikin2005}
I.~R. Shafarevich, Basic Notions of Algebra, Springer Berlin Heidelberg, 2005.
\newblock \href {https://doi.org/10.1007/b137643} {\path{doi:10.1007/b137643}}.

\bibitem{Skurichina1998}
M.~Skurichina, R.~P. Duin, Bagging for linear classifiers, Pattern Recognit.
  31~(7) (1998) 909--930.
\newblock \href {https://doi.org/10.1016/s0031-3203(97)00110-6}
  {\path{doi:10.1016/s0031-3203(97)00110-6}}.

\bibitem{platt1999probabilistic}
J.~Platt, et~al., Probabilistic outputs for support vector machines and
  comparisons to regularized likelihood methods, Advances in large margin
  classifiers 10~(3) (1999) 61--74.

\bibitem{Zadrozny2002}
B.~Zadrozny, C.~Elkan, Transforming classifier scores into accurate multiclass
  probability estimates, in: Proceedings of the eighth ACM SIGKDD international
  conference on Knowledge discovery and data mining - KDD '02, ACM Press, 2002,
  pp. 694--699.
\newblock \href {https://doi.org/10.1145/775047.775151}
  {\path{doi:10.1145/775047.775151}}.

\bibitem{Zadrozny2001}
B.~Zadrozny, C.~Elkan, Learning and making decisions when costs and
  probabilities are both unknown, Proceedings of the Seventh ACM SIGKDD
  International Conference on Knowledge Discovery and Data Mining (Jul. 2001).
\newblock \href {https://doi.org/10.1145/502512.502540}
  {\path{doi:10.1145/502512.502540}}.

\bibitem{ekin1999distance}
O.~Ekin, P.~L. Hammer, A.~Kogan, P.~Winter, Distance-based classification
  methods, INFOR: Information Systems and Operational Research 37~(3) (1999)
  337--352.
\newblock \href {https://doi.org/10.1080/03155986.1999.11732388}
  {\path{doi:10.1080/03155986.1999.11732388}}.

\bibitem{gomez2010committees}
V.~Gómez-Verdejo, J.~Arenas-García, A.~R. Figueiras-Vidal, Committees of
  adaboost ensembles with modified emphasis functions, Neurocomputing 73~(7-9)
  (2010) 1289--1292.
\newblock \href {https://doi.org/10.1016/j.neucom.2009.12.011}
  {\path{doi:10.1016/j.neucom.2009.12.011}}.

\bibitem{ahachad2017boosting}
A.~Ahachad, L.~Álvarez Pérez, A.~R. Figueiras-Vidal, Boosting ensembles with
  controlled emphasis intensity, Pattern Recognit. Lett. 88 (2017) 1--5.
\newblock \href {https://doi.org/10.1016/j.patrec.2017.01.009}
  {\path{doi:10.1016/j.patrec.2017.01.009}}.

\bibitem{Trajdos2019}
P.~Trajdos, R.~Burduk, Combination of linear classifiers using score function
  -- analysis of possible combination strategies, in: Advances in Intelligent
  Systems and Computing, Springer International Publishing, 2019, pp. 348--359.
\newblock \href {https://doi.org/10.1007/978-3-030-19738-4\_35}
  {\path{doi:10.1007/978-3-030-19738-4\_35}}.

\bibitem{Zhao2015}
X.~Zhao, Y.~Li, Q.~Zhao, Mahalanobis distance based on fuzzy clustering
  algorithm for image segmentation, Digital Signal Process. 43 (2015) 8--16.
\newblock \href {https://doi.org/10.1016/j.dsp.2015.04.009}
  {\path{doi:10.1016/j.dsp.2015.04.009}}.

\bibitem{McLachlan1992}
G.~J. McLachlan, Discriminant Analysis and Statistical Pattern Recognition,
  Wiley Series in Probability and Mathematical Statistics: Applied Probability
  and Statistics, John Wiley \& Sons, Inc., 1992, a Wiley-Interscience
  Publication.
\newblock \href {https://doi.org/10.1002/0471725293}
  {\path{doi:10.1002/0471725293}}.

\bibitem{gurney1997an}
K.~Gurney, An introduction to neural networks, Taylor \& Francis, London, 1997.
\newblock \href {https://doi.org/10.4324/9780203451519}
  {\path{doi:10.4324/9780203451519}}.

\bibitem{Kuncheva1998}
L.~Kuncheva, J.~Bezdek, Nearest prototype classification: {Clustering,} genetic
  algorithms, or random search?, IEEE Trans. Syst., Man, Cybern. C 28~(1)
  (1998) 160--164.
\newblock \href {https://doi.org/10.1109/5326.661099}
  {\path{doi:10.1109/5326.661099}}.

\bibitem{Hall2009}
M.~Hall, E.~Frank, G.~Holmes, B.~Pfahringer, P.~Reutemann, I.~H. Witten, The
  {WEKA} data mining software, SIGKDD Explor. Newsl. 11~(1) (2009) 10.
\newblock \href {https://doi.org/10.1145/1656274.1656278}
  {\path{doi:10.1145/1656274.1656278}}.

\bibitem{Hllermeier2010}
E.~Hüllermeier, J.~Fürnkranz, On predictive accuracy and risk minimization in
  pairwise label ranking, Journal of Computer and System Sciences 76~(1) (2010)
  49--62.
\newblock \href {https://doi.org/10.1016/j.jcss.2009.05.005}
  {\path{doi:10.1016/j.jcss.2009.05.005}}.

\bibitem{Banerjee1999}
M.~Banerjee, M.~Capozzoli, L.~McSweeney, D.~Sinha, Beyond kappa: {A} review of
  interrater agreement measures, Can. J. Statistics 27~(1) (1999) 3--23.
\newblock \href {https://doi.org/10.2307/3315487} {\path{doi:10.2307/3315487}}.

\bibitem{garcia2008extension}
S.~Garcia, F.~Herrera, An extension on``statistical comparisons of classifiers
  over multiple data sets''for all pairwise comparisons, Journal of Machine
  Learning Research 9 (2008) 2677--2694.

\bibitem{Bergmann1988}
B.~Bergmann, G.~Hommel, Improvements of general multiple test procedures for
  redundant systems of hypotheses, in: Multiple Hypothesenpr\"ufung / Multiple
  Hypotheses Testing, Springer Berlin Heidelberg, 1988, pp. 100--115.
\newblock \href {https://doi.org/10.1007/978-3-642-52307-6\_8}
  {\path{doi:10.1007/978-3-642-52307-6\_8}}.

\bibitem{Sokolova2009}
M.~Sokolova, G.~Lapalme, A systematic analysis of performance measures for
  classification tasks, Information Processing \& Management 45~(4) (2009)
  427--437.
\newblock \href {https://doi.org/10.1016/j.ipm.2009.03.002}
  {\path{doi:10.1016/j.ipm.2009.03.002}}.

\end{thebibliography}

\end{document}